\documentclass[runningheads]{llncs}
\usepackage[T1]{fontenc}
\usepackage{multirow}

\usepackage{svg}
\usepackage{todonotes}
\usepackage{xcolor}
\usepackage{graphicx}

\begin{document}
%

\title{From Threat Intelligence to Firewall Rules: Semantic Relations in Hybrid AI Agent and Expert System Architectures}
\titlerunning{Semantic Relations in Hybrid AI Agent and Expert System Architectures}
%
\author{Chiara Bonfanti\inst{1}\orcidID{0009-0007-8015-7786} \and
Davide Colaiacomo\inst{1}\orcidID{0009-0001-1761-5218} \and
Luca Cagliero\inst{1}\orcidID{0000-0002-7185-5247}\and
Cataldo Basile\inst{1}\orcidID{0000-0002-8016-1490}}

\authorrunning{C. Bonfanti et al.}
%
\institute{Department of Control and Computer Engineering, Politecnico di Torino, Italy
\email{\{name,surname\}@polito.it}}
\maketitle              
\begin{abstract}
Web security demands rapid response capabilities to evolving cyber threats. 
Agentic Artificial Intelligence (AI) promises automation, 
but the need for trustworthy security responses is of the utmost importance. 
This work investigates the role of semantic relations in extracting information 
for sensitive operational tasks, such as configuring security controls for mitigating threats. 
To this end, it proposes to leverage hypernym-hyponym textual relations to extract relevant information from Cyber Threat Intelligence (CTI) reports. By leveraging a neuro-symbolic 
approach, the multi-agent system
automatically generates CLIPS code for an expert system creating firewall rules to block malicious network traffic. Experimental results show the superior performance
of the hy\-per\-nym-hy\-po\-nym retrieval strategy compared to various baselines
and the 
higher effectiveness of the agentic approach in mitigating threats.
\end{abstract}
\section{Introduction}
\label{sec:intro}
Web applications, being publicly accessible, are common targets of cyberattacks. Defenders must secure them with complex and costly measures. At the same time, attackers are increasingly relying on AI-assisted tools for automated exploitation of vulnerabilities~\cite {anthropic}.
This creates a structural asymmetry in Cybersecurity~\cite{anderson2001economics}: attackers only need to target a few weaknesses, while defenders must protect the entire system. As a result, threats are countered with high delays~\cite{lazer2026surveyagenticaicybersecurity}.

Intrusion Detection Systems (IDSs) and Intrusion Prevention Systems (IPSs) are security tools designed to analyze network or host activities. They are frequently employed in the above scenario.
IDSs monitor activities to detect suspicious behaviour and, when appropriate, alert an operator or another system. 
IPSs perform the same function, but can automatically take action to stop or contain the suspicious activity~\cite{IDSvsIPS}.

Defending systems is challenging, as one must select and configure the most appropriate security controls. This process relies on incident reports and technical data on attacker behaviour~\cite{anderson2001economics}. 
To translate such data into practical defensive actions, the research community has been considering applying AI to understand the threat described in the report and its consequences for the protected system~\cite{IPS-LSTM}.
However, existing AI-based approaches struggle with the inherent complexity of automatically categorizing sensitive data, which is often exacerbated by severe data imbalance across different threats.

This paper aims to bridge this gap with the capabilities of Agentic AI and LLMs to capture the semantics of security events and align them with defensive strategies and secure code generation. 

The main contributions are threefold:

\begin{itemize}
\item First, it proposes a \textbf{new methodology for extracting semantically rich information from CTI reports based on established taxonomic relationships}, i.e., hypernyms and hyponyms. Inspired by their use to drive semantic search across multiple natural language processing tasks, such as textual entailment recognition, search query expansion, and automated machine translation~\cite{hyponim_prompt}, we propose to leverage hypernyms and hyponyms to enhance the semantic understanding of security events and facilitate mapping to defensive strategies and secure code generation. 
\item Secondly, it presents an \textbf{AI-based agentic system} that (\textit{i}) maps extracted info to existing defensive measures and (\textit{ii}) generates CLIPS code artifacts that can be used to configure security controls.
The system analyses the threats described in input CTI reports, identifies security controls that can mitigate them, and generates appropriate filtering rules. 
\item Finally, we perform an \textbf{in-depth empirical evaluation} of both information extraction and code generation steps. Our method excels in semantic retrieval and code generation, achieving superior robustness on imbalanced data than baseline methods (approximately $7\%$ gain in F1 Score). 
\end{itemize}

The paper is organized as follows:
Section~\ref{sec:literature_review} reviews the literature and positions our work in the state of the art; 
 Section~\ref{sec:data} introduces the reference datasets; 
Section~\ref{sec:methodology} illustrates our methodology and contributions; 
Section~\ref{sec:res} presents the validation of our approach.
Finally, Section~\ref{sec:conclusions} discusses conclusions and future works.

\section{Related work}
\label{sec:literature_review}
Semantic extraction is at the core of the present work~\cite{semantic_extraction}.
Despite it having recently re-gained interest in the natural language processing community~\cite{meconi-etal-2025-large},
its relevance to the cybersecurity applications related to intrusion prevention is still limited. 
Our approach utilizes an iterative prompting strategy~\cite{cot}  to retrieve semantically meaningful features. This strategy employs pre-existing relationships in the input text to form the agent's knowledge base. Taxonomic relationships, such as hyponymy and hyperonymy, are well known in the literature~\cite{hyponim_review} and are a pillar of our LLM's usage~\cite{hyponim_prompt}. 
Unlike existing solutions that require additional training data to incorporate ontological structure~\cite{consist_NER}, ours is inference-based.

Our contribution (\emph{i}) leverages this information to solve the aforementioned task. Rather than directly mapping raw text to classifications or code generation, we prompt LLMs in three stages: first, extracting specific domain entities; then, abstracting them into semantic categories; and finally, performing specific operations based on these enriched representations. Unlike methods that perform extraction in a single pass, our method uses semantic information at each stage, enabling the model to leverage explicit structural knowledge to guide categorization and classification toward a progressively enriched understanding of the domain. This is beneficial in Cybersecurity, as CTI reports are often noisy and verbose, while defenders benefit from structured information.

The present strategy (\emph{ii}) extends the CoALA framework~\cite{coala_framework} with theories drawn from cognitive psychology. Whilst psychology-based AI contributions are well established~\cite{sarangi-etal-2025-agentic,ebbinghaus_memory_LLM,ebbinghaus_agent}, our focus is on the practical integration of Ebbinghaus's memory decay and Collins and Quillian's semantic network within a code-generation agent.
The Cybersecurity domain remains one of the most productive grounds for symbolic AI, as deterministic reasoning over structured knowledge is preferred to probabilistic methods. CLIPS has a long history in this domain~\cite{exp_sys_cyber,IDS_clips}, providing a rule-based forward-chaining inference engine.
Graph-based knowledge representation is well established in agent architectures~\cite{kb_graph_agent} and offers fertile ground for optimization~\cite{kb_graph_jurix}. Our agent extends the standard implementation with a graph-based knowledge base that tracks concepts and is beneficial to code generation; unlike the decision-tree approach in~\cite{CLIPS_updates_trees}, graphs support incremental updates more naturally.

\section{Data}
\label{sec:data}
CTI reports are analyst-written documents that gather empirical observations about cyber threats. They describe attackers' targets and \textit{modus operandi} through narrative and concrete evidence, such as malicious IP addresses or compromised URLs. They are often mapped to MITRE ATT\&CK (Adversarial Tactics, Techniques, and Common Knowledge), an open-source catalog of attack tactics and techniques. It provides a well-established vocabulary for mapping Cybersecurity prose to structured descriptions of attacker behaviours, thus enabling consistent comparisons across reports. Publicly available CTI-annotated datasets are highly skewed. This well-known characteristic of the Cybersecurity domain \cite{mdpi_annotation} creates a low-resource setting in which conventional AI methodologies can underperform.
\paragraph{\textbf{Dataset A.}} CTI Human-Annotated Labels (CTI-HAL) is a human-annotated CTI dataset built from 81 real reports, which contain statements manually mapped to MITRE ATT\&CK techniques (116 distinct in total); annotation reliability is assessed through inter-annotator agreement~\cite{dataset_ctiA}.

\paragraph{\textbf{Dataset B.}}
This corpus is made of CTIs gathered from the Center for Internet Security. It contains 66 malware entries, each including an analyst synopsis and network artifacts. The target labels are entities known as \textit{security capabilities}~\cite{capabilities}. 

\section{Methodology}
\label{sec:methodology}
\begin{figure}[htbp]
    \centering
\includegraphics[width=\textwidth]{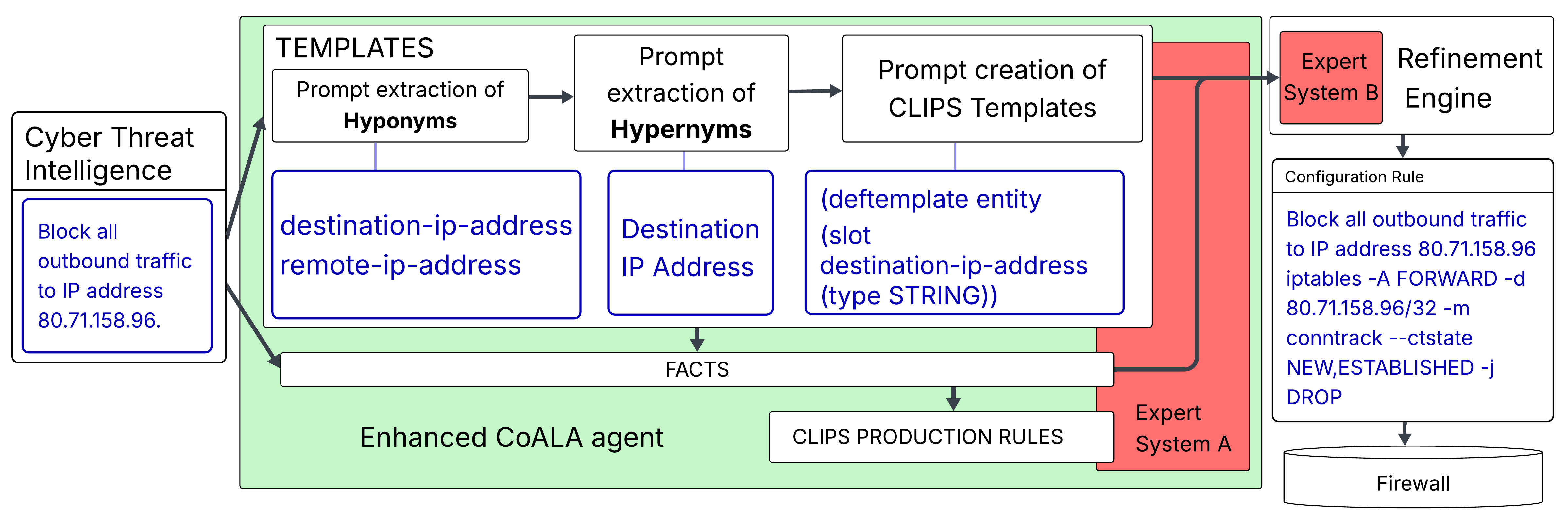}
\caption{The architecture of our agentic solution.}
    \label{fig:myimage}

\end{figure}

Our pipeline integrates neural and symbolic AI components, as shown in Fig.~\ref{fig:myimage}. It implements a Semantic Information Flow (SIF) that transforms CTI reports into filtering rules for security controls (iptables rules in Fig.~\ref{fig:myimage}). 
The SIF comprises neural components (green) and symbolic AI components (red). 

An Enhanced CoALA agent executes the initial steps of the SIF. It extracts semantic information from input reports by first retrieving hyponyms and then hypernyms of the security concepts expressed in the prose via an iterative call to an LLM. The last step uses the extracted hypernyms to build valid CLIPS templates that formally represent them.

The Expert System A role is of a syntactic verification layer against potential LLM hallucinations, already thinned out by guardrails and post-inference output controls, while the sequential instantiation of templates, facts, and production rules enables systematic detection of error propagation. 

The present study employs best practices to achieve deterministic LLM inference, thereby enabling the use of deterministic algorithms in PyTorch. The random seeds are fixed, CuDNN benchmarking is disabled, and eager attention is forced to avoid nondeterministic fused CUDA kernels. Greedy decoding and explicit device placement further reduced variability across runs. It is meant as a heuristic for determinism, and its application is thus beneficial.
CoALA architecture provides a sound foundation that has been extended to address known limitations in knowledge base manipulation, particularly given the constraints of CLIPS programming language {\tt deftemplate} structures. 

The Refinement Engine recognises the available security controls to counter threats described in CTI reports. Within this engine, Expert System B uses semantic information about the required security capabilities, leveraging the CLIPS rules generated by the agent. With this information, the engine produces the corresponding filtering rules. This final verification step ensures syntactical correctness, as inaccuracies would prevent acceptance by the security control.

\section{Results}
\label{sec:res}
The experimental evaluation was conducted using dedicated GPU hardware comprising NVIDIA RTX 4090 (44GB VRAM) and RTX 5090 (32GB VRAM) platforms, which provided sufficient computational resources to instantiate and compare multiple LLM architectures under controlled conditions. 

Two distinct tasks were defined: the first was to assess the effectiveness of the proposed semantic extraction method against existing solutions from the relevant literature; the second involved human annotators to provide an initial evaluation of the agentic framework's output (i.e., filtering rules for firewalls). 

\paragraph{\textbf{Task A}}
A multilabel classification experiment was conducted using Dataset A. Three different approaches were investigated: A-1, which employed static  (Word2Vec, GloVe) and contextualized embeddings (SecureBERT); A-2, which leveraged prompt-based methods (Chain-of-Thought and ours); A-3, which applied traditional machine learning methods, previously used in this domain~\cite{mdpi_annotation}. 
Qwen2.5-Coder-14B-Instruct was employed for A-2. LLM usage is tailored to Task A, as, to the best of our knowledge, this work is the first attempt to address semantic-enhanced prompting for CLIPS rule extraction. Evaluation metrics suited for imbalanced multilabel classification (e.g., Hamming loss) were used to ensure a fair assessment of model performance. Table~\ref{tab:resultsA} summarizes the results, with $k=10$ for the top-k accuracy scores.
Our semantic extraction excels in weighted F1 score and Top K Accuracy, demonstrating greater ability to shortlist relevant text snippets. BERTScore and ROUGE-L similarities confirm the pertinence of the selection, achieving maximal similarity in the longest common subsequence and competitive results at the embedding level. LLMs with Chain-of-Thought show comparable accuracy scores but significantly worse weighted F1 scores, mainly due to weaker performance on minority classes.

To deepen the analysis, we also explored Chain-of-Thought, Zero-Shot, and Few-Shot prompting augmented with semantic information. Each prompt included a confidence threshold, which guided the model's decisions.  The results are reported in Table~\ref{tab:final_results}. Hyponyms have been shown to be consistently more effective than hypernyms. Notably, enforcing weaker selectivity constraints (e.g., threshold=50\%) enhances performance.

\paragraph{\textbf{Task B}}The full pipeline 
described in Section~\ref{sec:methodology} was employed on Dataset B. 
Qwen2.5-Coder-14B-Instruct and Foundation-Sec-14B-Instruct were both considered; however, the latter's generative restrictions and limited CLIPS support made the former preferable. 
The evaluation, conducted by Cybersecurity experts, focused on qualitative judgment of our system's output. It involved multiple criteria, and the results in Table~\ref{tab:irr_metrics} show that all statistical scores used to evaluate annotators imply forms of alignment. Technical Correctness achieved the highest Krippendorff's alpha, indicating strong agreement on syntactic correctness. Scope Calibration, with the highest Spearman correlation, supports consistency in the relative ranking of rule scope. Fidelity to CTI remained high across all scores, indicating satisfactory agreement among annotators.

\begin{table}[htbp]
\centering
\small
\caption{Task A results comparing our hypernym-based prompting approach. Best-performer per metrics is written in boldface.}
\label{tab:resultsA}
\begin{tabular}{|l|l|l|l|l|l|l|}
\hline
\multirow{2}{*}{\textbf{Task}} & \multirow{2}{*}{\textbf{Method}} & \multicolumn{3}{c|}{\textbf{Task-based Eval.}} & \multicolumn{2}{c|}{\textbf{Semantic Metrics}} \\
\cline{3-7}
& & F1 w & Acc. w & Top-k Acc. & BERT Score F1 & ROUGE L \\
\hline

A-1 & Word2Vec & 0.070 & 0.990 & 0.250 & -- & -- \\
A-1 & GloVe & 0.118 & \textbf{0.991} & 0.382 & -- & -- \\
A-1 & SecureBERT & 0.043 & 0.990 & 0.210 & 0.831 & -- \\
A-2 & CoT & 0.308 & 0.932 & 0.935 & 0.862 & 0.402 \\
\textit{A-2} & \textit{Ours} & \textbf{\textit{0.329}} & \textit{0.934} & \textbf{\textit{0.968}} & \textit{0.858} & \textbf{\textit{0.444}} \\
A-3 & NB + OneVsRest & 0.122 & 0.911 & 0.947 & 0.900 & 0.337 \\
A-3 & SVM + OneVsRest & 0.062 & 0.910 & 0.947 & 0.845 & 0.261 \\
A-3 & SecureBERT + RF & 0.143 & 0.914 & 0.947 & 0.883 & 0.308 \\
A-3 & Binary Relevance RF & 0.183 & 0.916 & 0.842 & 0.885 & 0.290 \\
A-3 & Label Powerset SVM & 0.263 & 0.881 & 0.895 & \textbf{0.921} & 0.369 \\
\hline
\end{tabular}
\end{table}
\begin{table}[htbp]
\centering
\small
\caption{Task B Inter-rater reliability metrics by dimension.}
\label{tab:irr_metrics}
\resizebox{\columnwidth}{!}{%
\begin{tabular}{|l|c|c|c|c|c|}
\hline
\multirow{2}{*}{\textbf{Dimension}}
  & \multirow{2}{*}{\textbf{Krippendorff's} $\alpha$}
  & \multicolumn{3}{c|}{\textbf{Cohen's} $\kappa$}
  & \multirow{2}{*}{\textbf{Spearman's} $\rho$} \\
\cline{3-5}
  & & Unweighted & Linear & Quadratic & \\
\hline
Technical Correctness & $+0.5768$ & $+0.5371$ & $+0.5625$ & $+0.5817$ & $+0.6942$ \\
Fidelity to CTI       & $+0.5215$ & $+0.3445$ & $+0.4595$ & $+0.5599$ & $+0.5515$ \\
Scope Calibration     & $+0.5030$ & $+0.5827$ & $+0.6540$ & $+0.7251$ & $+0.7143$ \\
\hline
\end{tabular}
}
\end{table}
\begin{table}[h]
\small
\centering
\caption{Task A-2 prompting focused analysis. }
\label{tab:final_results}
\resizebox{\columnwidth}{!}{%
\begin{tabular}{|l|c|c|c|c|c|}
\hline
\textbf{Method} & \textbf{F1} & \textbf{ACC} & \textbf{TOP-10} & \textbf{BERTScore} & \textbf{ROUGE-L F1} \\
\hline
Our Method (3-Stage 50\%)  & \textbf{0.329} & 0.934 & 0.968 & 0.858 & \textbf{0.444} \\
Few-Shot + Hyponyms        & 0.311 & 0.935 & 0.952 & 0.860 & 0.411 \\
CoT + Hypernyms            & 0.308 & 0.933 & 0.984 & 0.865 & 0.407 \\
CoT Baseline               & 0.308 & 0.932 & 0.935 & 0.862 & 0.402 \\
Few-Shot + Hypernyms       & 0.287 & 0.929 & 0.984 & 0.856 & 0.400 \\
CoT + Hyponyms             & 0.294 & 0.931 & 0.984 & \textbf{0.871} & 0.391 \\
60\% Threshold             & 0.288 & 0.934 & \textbf{1.000} & 0.875 & 0.401 \\
Zero-Shot + Hyponyms       & 0.238 & 0.933 & 0.919 & 0.848 & 0.344 \\
Zero-Shot + Hypernyms      & 0.233 & 0.931 & 0.871 & 0.841 & 0.327 \\
Baseline (only semantic information)              & 0.173 & \textbf{0.937} & 0.774 & 0.843 & 0.270 \\
\hline
\end{tabular}
}
\end{table}

\section{Conclusions and future work}
\label{sec:conclusions}
This paper addresses the use of AI in Intrusion Prevention Systems. 
It explored the adoption of semantic extractors from CTI reports, where 
CLIPS code is generated for an expert system that creates firewall rules to block malicious network traffic.
 
As demonstrated by the results of Task A, the prompting methodology employed in this study 
based on hypernyms and hyponyms 
performed significantly better 
compared to baseline methods. 
This underscores the need for ad hoc semantic retrieval modules rather than classical classification approaches used for IDS. 

Although the current agent system is intended for research purposes, the findings from Task B show promising results for future usability and deployability, as human annotators achieved satisfactory inter-annotator agreement. This leaves room for further development for downstream usage of filtering rules.

Future studies should also focus on exploring determinism in LLM use. In this article, it is a heuristic for true determinism. Additional investigation in this direction is recommended. 

The present and future works sought to take steps towards a more reliable use of Large Language Models in sensitive domains, such as cybersecurity.

%
%
\bibliographystyle{splncs04}
\bibliography{bibliography}
\end{document}